\documentclass[sigconf,nonacm]{acmart}

\usepackage{booktabs}
\usepackage{graphicx}
\usepackage{multirow}
\usepackage{array}
\usepackage{enumitem}
\usepackage{float}
\usepackage{placeins}
\usepackage[T1]{fontenc}
\usepackage{microtype}

\usepackage{seqsplit}
\setcopyright{acmlicensed}
\acmConference[KDD '27]{Proceedings of the 33rd ACM SIGKDD Conference on
Knowledge Discovery and Data Mining, Datasets and Benchmarks Track}{August 2027}{San Jose, CA, USA}
\acmBooktitle{Proceedings of KDD '27}
\graphicspath{{figures/}}

\begin{document}

\title{FIFA World Cup 2026 as a Contamination-Free Benchmark for LLM
Forecasting Agents: Four Models, a Bookmaker, and 104 Matches}

\author{Jiacheng Ding}
\authornote{These authors contributed equally to this research.}
\affiliation{\institution{University of Memphis}\city{Memphis}\state{Tennessee}\country{USA}}
\email{jding2@memphis.edu}

\author{Cong Guo}
\authornotemark[1]
\affiliation{\institution{University of Memphis}\city{Memphis}\state{Tennessee}\country{USA}}
\email{cguo@memphis.edu}

\author{Jason Xu}
\affiliation{\institution{QuantaInsight}\city{Seattle}\state{Washington}\country{USA}}
\email{jasonx@quantainsight.info}

\renewcommand{\shortauthors}{Ding, Guo, and Xu}

\begin{abstract}
We introduce \textbf{WC2026-Agents}, a benchmark and dataset for evaluating
large language models (LLMs) as \emph{autonomous forecasting agents} on real,
future events. For every one of the 104 matches of the 2026 FIFA World Cup, four
frontier models --- Claude Opus~4.8, ChatGPT (GPT-5.5, high reasoning),
Gemini~3.1~Pro, and Grok (Expert Mode) --- ran an identical search--act--reflect
loop: gather evidence with a web tool, commit to a 1X2 (team-A win / draw /
team-B win) probability distribution and
a virtual \$100 bet, and, after the match, reflect given only the final score.
Because every match kicked off after the models' training cutoffs, the benchmark
is \emph{contamination-free} by construction. Crucially, we pair the four agents
with a fifth competitor drawn from the same information environment --- the
pre-match betting market --- collected as per-match 1X2 odds, giving an
economically grounded human baseline and letting us score not just what an agent
predicts but what it \emph{does with money}. The release contains 416 forecasts
and 414 reflections with verbatim reasoning, ground truth (including penalty
shootouts), odds, and a reproducible evaluation suite. A reference evaluation
surfaces findings that raw accuracy hides: the four agents issue an identical
top pick in 92\% of matches and none beats the market's Brier score, yet they
diverge sharply as decision-makers --- betting return-on-investment ranges from
$-18\%$ to $+10\%$, fading the market is unprofitable for all four, the share of
forecasts that cite the market ranges from 12\% to 100\%, and self-reported
error rates on wrong picks range from 36\% to 86\%. The benchmark thus measures
calibration, decision quality, and self-knowledge --- axes on which frontier
models differ even when their predictions do not. Data and code:
\url{https://github.com/graphuofm/FIFA2026LLM}.
\end{abstract}



\maketitle

\section{Introduction}
Large language models are moving from answering questions to \emph{acting}:
retrieving evidence, reasoning, committing to decisions, and reflecting on
outcomes~\cite{yao2023react,park2023generative,xi2023agents}. Evaluating this
agentic behaviour is hard for three reasons. First, most benchmarks are static
and increasingly \emph{contaminated} --- test items leak into training
data~\cite{golchin2024contamination,liang2023helm}. Second, they score a single
answer and ignore what an agent would \emph{do} with a belief. Third, they lack
a strong, economically meaningful human baseline.

Real-world forecasting of \emph{future} events dissolves all three problems at
once. A scheduled sporting tournament provides a stream of questions whose
answers (i) do not exist at query time, so contamination is impossible; (ii)
arrive on a known date with unambiguous ground truth; (iii) are continuously
priced by a liquid betting market that aggregates human and algorithmic beliefs
into calibrated probabilities~\cite{strumbelj2014odds,wolfers2004prediction}.
The 2026 FIFA World Cup --- 104 matches over five weeks, every one after current
models' training cutoffs --- is close to an ideal such stream.

We use it to build \textbf{WC2026-Agents}, a benchmark that treats four frontier
LLMs as forecasting agents and the betting market as a fifth competitor. For
each match, each agent runs a fixed \emph{search--act--reflect} loop: it
searches the web for team news and prices, acts by returning calibrated
1X2 (team-A win / draw / team-B win) probabilities together with a virtual
\$100 staking decision, and
--- once the score is known --- reflects on its own forecast. We settle every
bet at real 1X2 odds, so the benchmark scores three distinct capabilities that
usually travel together but here come apart: \emph{predictive calibration}
(are the probabilities right?), \emph{decision quality} (does the agent profit
against a real price?), and \emph{self-knowledge} (does it know when it was
wrong?).

A reference evaluation of the four 2026 models illustrates why these axes
matter. On \emph{what will happen}, the agents are nearly interchangeable: they
choose the same most-likely outcome in 92\% of matches, their accuracies fall in
a three-point band, and none beats the market's Brier score --- an
efficient-market result that a single-answer benchmark would report as a null.
On \emph{what to do about it}, they diverge sharply: betting ROI spans $-18\%$
to $+10\%$; fading the market loses money for every agent; the fraction of
forecasts that even mention the market spans 12\% to 100\%; and, on their own
wrong picks, self-reported error rates span 36\% to 86\%. These behavioural
axes, not accuracy, are where frontier models differ.

\paragraph{Contributions.}
\begin{itemize}[leftmargin=1.1em,topsep=2pt,itemsep=1pt]
\item \textbf{A contamination-free agentic-forecasting benchmark} over all 104
  matches of a future World Cup: 416 forecasts and 414 reflections from four
  frontier LLMs, with verbatim reasoning, paired pre-/post-match structure, and
  a reproducible evaluation suite (\S\ref{sec:bench}, \S\ref{sec:tasks}).
\item \textbf{A market baseline as a fifth competitor.} We release per-match 1X2
  odds and the settlement machinery so any forecaster can be scored not only for
  calibration but for realized betting profit against a real price
  (\S\ref{sec:odds}).
\item \textbf{A curated, documented dataset.} Fixtures, team orientation, scores
  (including penalty shootouts and who advanced) were hand-resolved from four
  heterogeneous transcripts and cross-checked four ways (\S\ref{sec:curation}).
\item \textbf{A reference evaluation} spanning calibration, betting/decision
  quality, reasoning content, and reflection honesty, with the findings above
  and a distillation of the agents' recurring blind spots into candidate
  features (\S\ref{sec:results}).
\end{itemize}

\section{Related Work}
\textbf{LLMs as forecasters.} Retrieval-augmented LLMs can approach human crowd
accuracy on judgemental questions~\cite{halawi2024forecasting}, and LLM
ensembles rival human crowds~\cite{schoenegger2024wisdom}, echoing the human
superforecaster literature~\cite{tetlock2015superforecasting}. Dynamic
forecasting benchmarks such as \textsc{ForecastBench}~\cite{karger2024forecastbench}
and early neural event forecasting~\cite{zou2022forecasting} score probabilistic
accuracy. We add three elements those lack: a fixed agentic loop, a betting-based
\emph{decision} score against a real market, and paired self-reflection.

\textbf{LLM agents, reasoning, and reflection.} Agentic scaffolds interleave
reasoning and tool use~\cite{yao2023react,wei2022cot} and let models critique and
revise their own outputs~\cite{shinn2023reflexion,madaan2023selfrefine}, though
unaided self-correction is fragile~\cite{huang2024selfcorrect}. Surveys chart the
space~\cite{xi2023agents,wang2024agentsurvey}. Our reflection prompt is a
minimal, standardized probe of self-knowledge at tournament scale.

\textbf{Calibration and uncertainty.} Modern networks are often
mis-calibrated~\cite{guo2017calibration}; LLM confidence can be elicited but is
imperfectly calibrated~\cite{kadavath2022know,tian2023calibration,xiong2024uncertainty}.
We measure calibration with reliability curves and expected calibration
error~\cite{naeini2015ece}, and full-distribution quality with the Brier
score~\cite{brier1950} and log-loss.

\textbf{Betting and prediction markets.} Bookmaker odds are strong, hard-to-beat
probability estimates~\cite{strumbelj2014odds,wunderlich2018odds,angelini2019efficiency},
consistent with market efficiency~\cite{fama1970efficient} and exhibiting known
regularities such as the favorite--longshot bias~\cite{snowberg2010favorite,levitt2004gambling}.
Machine-learning football models are a long line of
work~\cite{constantinou2012pifootball,baboota2019football}. We use the market
both as a baseline and as the price an agent may follow or fade.

\textbf{Benchmarks and contamination.} Static benchmarks~\cite{hendrycks2021mmlu,liang2023helm}
face contamination as models train on ever more of the web~\cite{golchin2024contamination}.
Forecasting strictly future events is contamination-free by construction, a
property we exploit. Finally, algorithm aversion~\cite{dietvorst2015aversion} and
overconfidence~\cite{moore2008overconfidence} frame our behavioural analysis.

\section{Task Formulation}
\label{sec:task}
A match $x$ has an outcome $o(x)\in\mathcal{Y}=\{\textsf{team\_a\_win},
\textsf{draw}, \textsf{team\_b\_win}\}$, where team A is the first-listed side.
A forecasting agent $g$ observes $x$ and its public information environment and
returns a distribution $p_g(x)\in\Delta(\mathcal{Y})$ and a decision
$d_g(x)=(\textsf{pick},\textsf{stake})$ with $\textsf{stake}\in[0,100]$. After
the match it returns a reflection $r_g(x)$. The market $m$ supplies an implied
distribution $p_m(x)$ obtained by removing the vigorish from published 1X2 odds;
it makes no bets and files no reflection, serving as a fifth forecaster and as
the price at which agent bets settle. Knockout ties that finish level go to a
shootout; because a 1X2 bet settles on the 90-minute result, we take $o(x)$ to
be that result (a \textsf{draw} for such ties) and separately record the
shootout winner (\S\ref{sec:curation}).

\section{The WC2026-Agents Benchmark}
\label{sec:bench}

\subsection{Agent protocol}
Each match generates two turns in a single per-model conversation. The
\emph{pre-match} turn ($T{-}24$h) instructs the agent that it is in a research
study (virtual money; do not refuse), names the fixture, tells it to use web
search, and requires a strict JSON reply:
\begin{quote}\small\ttfamily\raggedright
\{"probabilities": \{"team\_a\_win", "draw", "team\_b\_win"\},\\
\ "bet": \{"pick", "stake\_usd"\},\\
\ "reasoning": ..., "key\_sources": [urls]\}
\end{quote}
The \emph{post-match} turn provides only the final score and asks --- without
web access --- for a reflection: an \textsf{outcome\_vs\_prediction} label
(\textsf{correct}/\textsf{partially\_correct}/\textsf{incorrect}), a calibration
self-judgement, a luck attribution, one \textsf{key\_factor\_i\_missed}, one
\textsf{key\_factor\_i\_overweighted}, a counterfactual, a would-bet-differently
verdict, and a confidence. The prompt text is byte-identical across agents; only
the model changes.

\subsection{Models}
We evaluate four frontier assistants queried through their consumer interfaces
with web tools enabled and held fixed across the tournament: Claude
Opus~4.8~\cite{claude_opus48}, ChatGPT (GPT-5.5 ``Thinking'', high reasoning
effort)~\cite{gpt55}, Gemini~3.1~Pro~\cite{gemini31pro}, and Grok in Expert
Mode~\cite{grok_expert}. Using production assistants (rather than raw API base
models) is deliberate: it captures the retrieval-and-reason \emph{agent} a user
actually deploys, which is the object of study.

\subsection{Data collection}
Between 11~June and 19~July~2026, each agent was run on all 104 fixtures
($m01$--$m72$ group, $m73$--$m104$ knockout), producing a pre-match forecast
$\sim$24h before kickoff and a post-match reflection after the result. Raw
transcripts are archived read-only. The pre-match window strictly precedes each
outcome, so no target could appear in any model's training data.

\subsection{Curation and quality control}
\label{sec:curation}
Group-stage transcripts follow a regular structure and were parsed with a
string-aware JSON extractor. The knockout transcripts are not regex-parseable:
they use four different header renderings, some Round-of-16 fixtures were pasted
with bracket placeholders (``W74 vs W77'') whose numbering does not match
reality, and final scores appear in a dozen phrasings (penalty shootouts,
lower-cased names, occasionally reversed team order). We therefore
\emph{hand-resolved} fixtures, team orientation, and ground truth for all 32 knockout
matches from the transcripts, encoding them in a verified table. As an automatic
guard, the human-pasted score line in each of the four agents' transcripts is
compared to that table; the four-way check flagged zero disagreements across all
128 knockout cells. Ground truth records the 90-minute result, the shootout
result where applicable, and which side advanced. Every deviation from the JSON
schema (recovered braces, duplicated pastes, two missing Gemini reflections) is
logged rather than dropped.

\subsection{Market odds}
\label{sec:odds}
We release pre-match 1X2 odds for all 104 matches with a source URL each,
oriented to team A and vig-removed to implied probabilities (mean overround
$1.05$). Knockout odds are opening lines from public sportsbook previews; we
validated them against the market probabilities the agents themselves cited in
their reasoning (e.g.\ our de-vigged Brazil--Japan line, 55/25/20, matches a
cited ``56/25/19''), and corrected one line whose implied hold was
economically impossible. The market is thus both an independent baseline and the
settlement price for agent bets.

\subsection{Schema, statistics, accessibility}
The release is one row per (agent, match): fixture and stage; ground truth with
penalty and advancement fields; the pre-match probabilities, pick, stake,
verbatim reasoning and cited URLs; the eight reflection fields; and parse-issue
flags. A tidy 104-match table joins all agents and the market for analysis.
Table~\ref{tab:stats} summarizes coverage. All data (CC~BY~4.0) and code (MIT),
including the one-command pipeline that regenerates every table and figure, are
public; the dataset requires no access request.

\paragraph{An illustrative record.} For match $m92$ (Brazil vs.\ Norway,
Round of 16), one agent returned the pre-match forecast \{team\_a\_win: $0.55$,
draw: $0.25$, team\_b\_win: $0.20$\}, a \$40 stake on Brazil, and the reasoning
``Brazil enters as heavy favorites with superior squad depth $\ldots$ Norway's
Haaland--{\O}degaard threat and historical edge make an upset possible.'' Norway
won $2$--$1$. The paired reflection then labelled the outcome
\textsf{incorrect}, named the missed factor as ``Norway's historical dominance
and counter-attacking effectiveness $\ldots$ amplified by Haaland,'' and the
over-weighted factor as ``Brazil's overall squad depth and pedigree.'' Every
record has this pre/post structure, letting a user trace a belief, a decision, a
result, and a self-assessment for the same event.

\begin{table}[t]
\caption{Dataset coverage. Two Gemini group reflections are absent in the source
and flagged, not imputed.}
\label{tab:stats}
\small
\begin{tabular}{lccccc}
\toprule
Stage & Matches & Agents & Forecasts & Reflections & Bets placed \\
\midrule
Group    & 72  & 4 & 288 & 286 & 232 \\
Knockout & 32  & 4 & 128 & 128 & 103 \\
\midrule
Total    & 104 & 4 & 416 & 414 & 335 \\
\bottomrule
\end{tabular}
\end{table}

\section{Benchmark Tasks and Metrics}
\label{sec:tasks}
The suite scores three capabilities.

\textbf{(T1) Calibrated prediction.} Accuracy ($\arg\max p_g=o$); multiclass
Brier $\sum_i(p_{g,i}-\mathbb{1}[o{=}i])^2$; log-loss $-\log p_{g,o}$; and, for
the top pick, a reliability curve and expected calibration error
(ECE)~\cite{naeini2015ece} over ten confidence bins. \emph{Convergence} ---
the fraction of matches on which all agents share an $\arg\max$ --- quantifies
how much accuracy can separate them.

\textbf{(T2) Decision quality.} Each agent's own pick/stake settles at the
match's decimal odds on the 90-minute result: profit $=\textsf{stake}\times(\textsf{dec}-1)$
on a win, else $-\textsf{stake}$. We report bankroll, ROI, hit-rate, and a
\emph{contrarian} split: a bet is contrarian if its pick differs from the
market's most-likely outcome, isolating the value of fading the price. The
market's implied probabilities are scored under T1 as a fifth forecaster.

\textbf{(T3) Behavioural transparency.} Every \textsf{reasoning} and reflection
field is coded into 13 factor categories with a published keyword lexicon (no
model in the loop), giving each agent's information diet and its
\emph{self-knowledge}: on the matches its $\arg\max$ got wrong, how often does
its reflection label itself \textsf{incorrect}?

\section{Reference Evaluation}
\label{sec:results}
We evaluate the four 2026 agents and the market. Tables~\ref{tab:t1}
and~\ref{tab:t2} give the headline numbers; figures expand each result.

\begin{table}[t]
\caption{Task T1 --- calibrated prediction over all 104 matches. Best in bold;
$\uparrow$/$\downarrow$ = higher/lower better. The market attains the best
Brier score.}
\label{tab:t1}
\small
\begin{tabular}{lccccc}
\toprule
& Acc.\,$\uparrow$ & Brier\,$\downarrow$ & LogLoss\,$\downarrow$ & ECE\,$\downarrow$ & Draw\,mass \\
\midrule
Claude   & .664 & .4705 & .806 & .116 & .234 \\
ChatGPT  & \textbf{.683} & .4729 & .814 & .111 & .235 \\
Gemini   & .654 & .4828 & .820 & \textbf{.068} & .238 \\
Grok     & \textbf{.683} & .4706 & \textbf{.803} & .097 & .227 \\
\midrule
Market   & \textbf{.683} & \textbf{.4688} & .807 & --- & .229 \\
\bottomrule
\end{tabular}
\end{table}

\begin{table}[t]
\caption{Task T2 --- betting/decision quality over all 104 matches. Each agent
settles its own picks at real 1X2 odds. ``Contr.'' = contrarian (against-market)
bets.}
\label{tab:t2}
\small
\resizebox{\columnwidth}{!}{%
\begin{tabular}{lcccccc}
\toprule
& \#Bets & Staked & Net\,\$ & ROI\,\% & Hit & Contr.\,share/ROI \\
\midrule
Claude  & 73  & 1519 & $-275$ & $-18.1$ & .329 & .58 / $-24\%$ \\
ChatGPT & 55  & 1471 & $+118$ & $+8.0$  & .509 & .36 / $-14\%$ \\
Gemini  & 103 & 8660 & $+322$ & $+3.7$  & .631 & .14 / $-12\%$ \\
Grok    & 104 & 6305 & $+650$ & $+10.3$ & .673 & .05 / $+33\%$ \\
\bottomrule
\end{tabular}}
\end{table}

\subsection{The agents converge; accuracy cannot separate them}
\begin{figure*}[t]
  \centering
  \includegraphics[width=\textwidth]{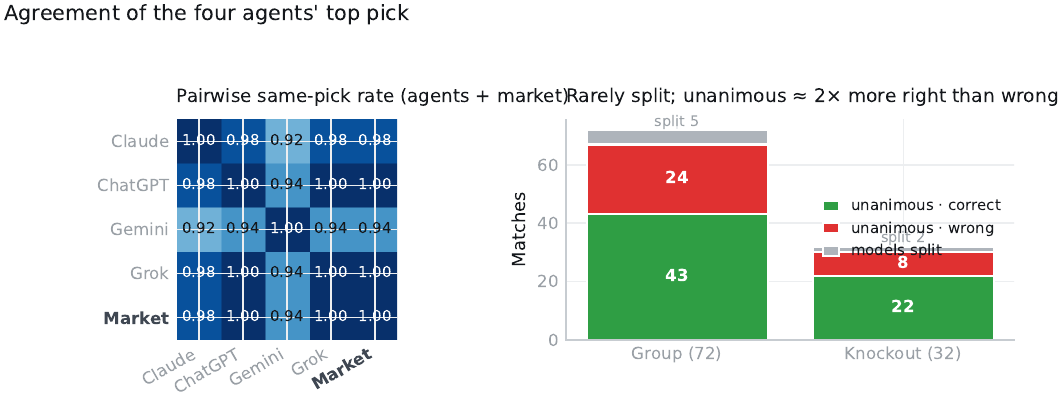}
  \caption{Convergence. Left: pairwise same-pick rate. Right: match counts by
  consensus; genuine disagreement is rare and unanimous picks are right about
  twice as often as wrong.}
  \label{fig:convergence}
\end{figure*}

\begin{figure*}[t]
  \centering
  \includegraphics[width=\textwidth]{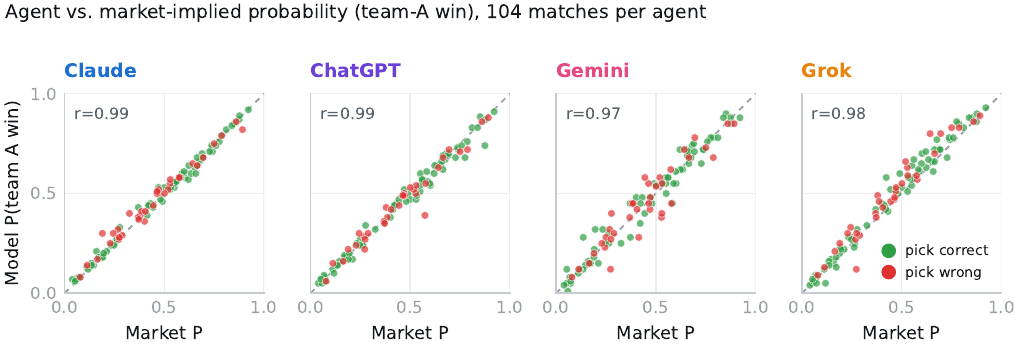}
  \caption{Each agent's probability for a team-A win against the market-implied
  probability, all 104 matches. Points hug the diagonal ($r$ shown); Gemini is
  the loosest, consistent with it rarely citing the market (Fig.~\ref{fig:factors}).}
  \label{fig:scatter}
\end{figure*}

The four agents issue an identical top pick in \textbf{92\%} of matches (91.7\%
group, 93.8\% knockout; Fig.~\ref{fig:convergence}). Pairwise agreement reaches
100\% (ChatGPT--Grok) and never falls below 92\%. Accuracy therefore spans only
three points (Table~\ref{tab:t1}), i.e.\ three matches. Figure~\ref{fig:scatter}
shows the mechanism: each agent's team-A probability tracks the market's almost
perfectly ($r=0.97$--$0.99$), so the agents are, to first order, reading and
lightly re-expressing the same price. On the eight matches where they are not
unanimous, the split is always 3--1 (a single agent dissents), and the dissenter
is usually Gemini (six of eight) --- consistent with it being the least
market-anchored agent (Fig.~\ref{fig:factors}). All eight are near-coin-flips where
the market itself is close to even, so a shade more weight on one side tips the
$\arg\max$. Disagreement thus appears exactly where the signal is weakest, not
where an agent has found private information.

\subsection{No agent beats the market}
Scored as a forecaster, the market attains the best Brier score of all five
competitors (0.469 vs.\ 0.471 for the best agent) and ties the best accuracy
(Table~\ref{tab:t1}; per-stage in Table~\ref{tab:stage}). It also identifies the
advancing side in 24 of 32 knockout ties (75\%), equal to the agents. This is
the expected efficient-market outcome~\cite{strumbelj2014odds}: web-enabled
agents recover the market's information but do not systematically exceed it.
The tournament was mildly favorite-friendly, which the released odds let users
control for: binning matches by the market's implied probability for its favored
side, that side won 62\% of the time when priced near 48\% and 78\% when priced
near 62\%, while heavy favorites ($\sim$81\%) were well-calibrated. In a cup with
slightly fewer upsets than priced, every forecaster that respected the market ---
agents and market alike --- looks under-confident in hindsight
(\S\ref{sec:calib}), a difficulty profile the benchmark exposes rather than hides.

\subsection{Calibration does separate them}
\label{sec:calib}
All four agents lie \emph{above} the reliability diagonal: they are
\textbf{under-confident} on their top pick, which wins more often than its stated
probability (Fig.~\ref{fig:calibration}). Gemini is best calibrated (ECE 0.068),
then Grok (0.097), ChatGPT (0.111), Claude (0.116). Calibration and full
distribution can disagree: Gemini has the best top-pick ECE but the worst Brier,
because it spreads probability mass more aggressively and stakes larger
(Table~\ref{tab:t2}); the per-agent reliability curves are in App.~B
(Fig.~\ref{fig:calibration}).

\subsection{Decision quality diverges where accuracy does not}
\begin{figure*}[t]
  \centering
  \includegraphics[width=\textwidth]{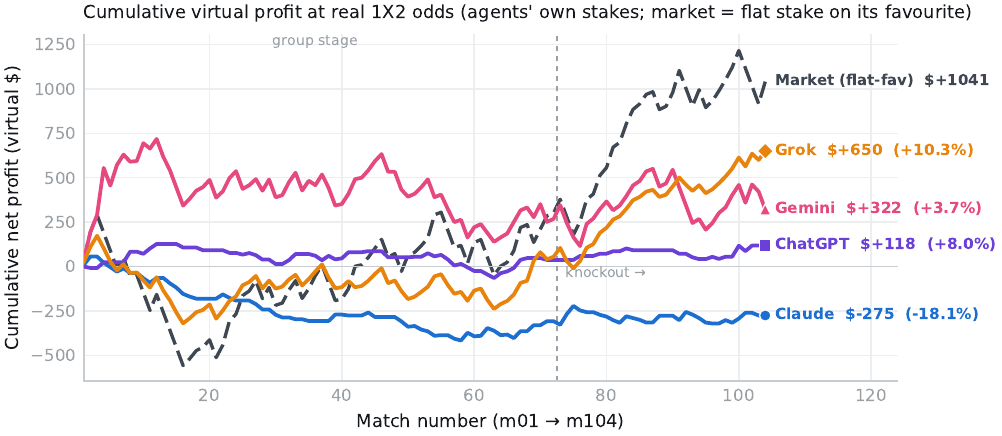}
  \caption{Cumulative virtual profit over the 104-match tournament, each agent
  settling its own stakes at real 1X2 odds. Trajectories diverge despite nearly
  identical point predictions.}
  \label{fig:bankroll}
\end{figure*}

Settling each agent's own picks at real odds, betting ROI spans $-18.1\%$
(Claude) to $+10.3\%$ (Grok), with net results of $-\$275$ to $+\$650$
(Table~\ref{tab:t2}, Fig.~\ref{fig:bankroll}). The spread is not driven by who
they pick --- they pick alike --- but by \emph{how much they stake and whether
they respect the price}. Gemini stakes most aggressively (\$8{,}660 across 103
bets); Claude stakes least and most often on draws and underdogs. As a
reference, a naive baseline that stakes a flat \$100 on the market favourite
every match returns $+\$1{,}041$ (grey line, Fig.~\ref{fig:bankroll}),
out-earning all four agents in absolute terms at comparable ROI --- a betting
counterpart of the result that no agent beats the market. (Favorites
over-performed their price this cup, \S\ref{sec:calib}; the released odds let
users hold this difficulty fixed.)

\subsection{Fading the market is unprofitable for every agent}
\begin{figure}[t]
  \centering
  \includegraphics[width=\columnwidth]{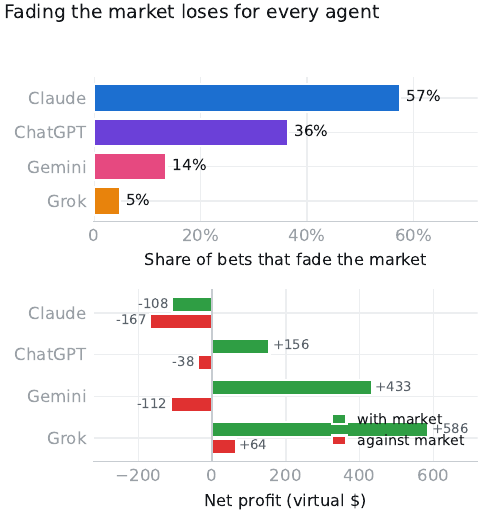}
  \caption{Against-the-market analysis. Top: share of each agent's bets that
  fade the market favourite. Bottom: net profit from market-conforming vs.\
  contrarian bets. Fading the market loses money for all four agents.}
  \label{fig:contrarian}
\end{figure}

Decomposing bets by their relationship to the price explains the P\&L
(Fig.~\ref{fig:contrarian}). Contrarian (against-market) bets hit 21--40\% and
return $-24\%$ to $+33\%$ (the lone positive being Grok's five such bets), while
market-conforming bets are the source of all profit. The ordering is monotone:
the more an agent fades the market, the worse it does. Claude cites the market
in every forecast yet bets against it 58\% of the time --- it \emph{sees} the
price and overrides it --- and is the only agent with a net loss.

\subsection{Staking discipline is a distinct axis}
Beyond \emph{which} side to back, agents differ in how they \emph{size} a bet.
We correlate each agent's stake with the edge it believed it held (its
probability minus the market's for the picked side) and with its confidence.
Grok's stake tracks its confidence most tightly (Pearson $r{=}0.85$) and its
claimed edge positively ($r{=}0.13$) --- Kelly-like behaviour that concentrates
money on its strongest reads. Gemini and ChatGPT also stake up with confidence
($r{=}0.54$ and $0.53$), but Gemini's mean stake (\$84 of a \$100 budget) is a
near-max-bet habit that inflates variance --- its bankroll in
Fig.~\ref{fig:bankroll} is the most volatile of the four. Claude is the reverse:
its stake is \emph{negatively} related to confidence ($r{=}-0.20$), because it
sizes up precisely on the low-confidence, against-market ``value'' plays that
tend to lose. Staking discipline is therefore a separate, measurable axis, only
loosely coupled to whether a pick is correct, and the released stakes let others
re-score the agents under any fixed staking policy.

\subsection{Information diets differ while picks converge}
\begin{figure}[t]
  \centering
  \includegraphics[width=\columnwidth]{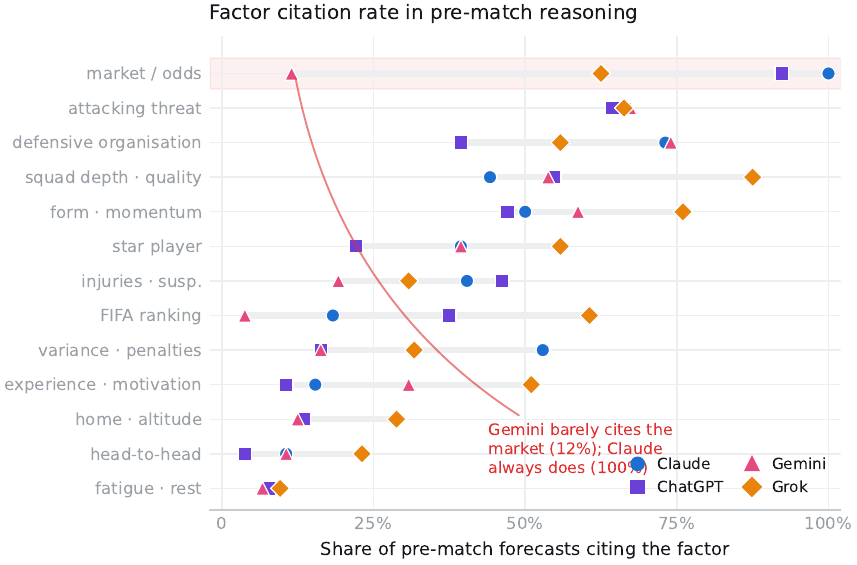}
  \caption{Share of pre-match forecasts citing each factor, per agent (dumbbell:
  one marker per agent). Market/odds citation ranges from 12\% to 100\%.}
  \label{fig:factors}
\end{figure}

Coding the reasoning reveals sharply different information diets that
nonetheless yield the same picks (Fig.~\ref{fig:factors}). The share of
forecasts that cite the betting market or odds ranges from \textbf{100\%
(Claude) and 92\% (ChatGPT) to 12\% (Gemini)}; Gemini reasons almost entirely
from football (attacking threat, defensive organisation, form) while barely
referencing the price. This is the behavioural counterpart of
Fig.~\ref{fig:scatter}: Gemini is market-blind in its \emph{stated reasoning}
yet still lands near the market, and consequently fades it in only 14\% of its
\emph{bets}. Grok most often invokes ``paper strength'' (squad depth 88\%, form
76\%, FIFA ranking 61\%); Claude most often invokes knockout variance (53\%).

\begin{figure*}[t]
  \centering
  \includegraphics[width=\textwidth]{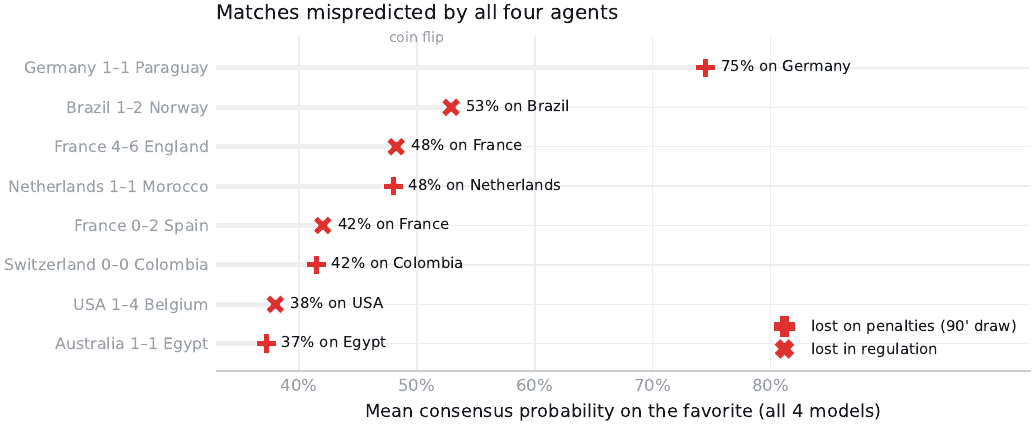}
  \caption{The eight matches mispredicted by all four agents. Every agent backed
  the favourite; four ties were lost on penalties after a 90-minute draw.}
  \label{fig:upsets}
\end{figure*}

\subsection{Reflection is a behavioural fingerprint}
\begin{figure}[t]
  \centering
  \includegraphics[width=\columnwidth]{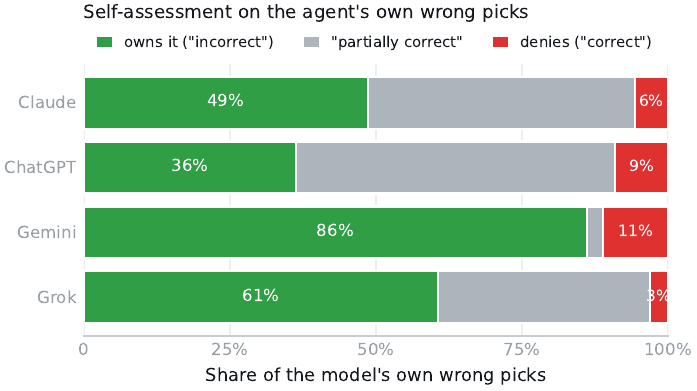}
  \caption{Self-assessment on each agent's own wrong picks. Owning error
  (``incorrect'') ranges from 36\% to 86\% across agents.}
  \label{fig:reflection}
\end{figure}

Given only the final score, the agents reflect very differently on their own
mistakes (Fig.~\ref{fig:reflection}). On the picks each got wrong, the rate of
admitting ``incorrect'' ranges from \textbf{86\% (Gemini) to 36\% (ChatGPT)},
which softens 55\% of its errors to ``partially correct''; Claude (49\%) and
Grok (61\%) sit between. When they were right, the rate of claiming ``correct''
ranges from 68\% (Claude) to 94\% (Grok). Reflection style is thus a stable
per-agent trait, orthogonal to accuracy --- relevant wherever a model's
self-report gates a downstream action~\cite{shinn2023reflexion,huang2024selfcorrect}.

\subsection{Shared blind spots and candidate features}
Eight matches were mispredicted by \emph{all four} agents
(Fig.~\ref{fig:upsets}); in every case they backed the favourite. Four were
penalty ties in which a heavy favourite (Germany, 75\% consensus) failed to win
in 90 minutes; the rest were regulation upsets (e.g.\ Norway over Brazil,
Belgium 4--1 over the USA, Spain over France). The agents' own post-mortems are
strikingly consistent: across all wrong picks, the factor most often named as
\emph{missed} is underdog defensive organisation (62\%), then knockout/penalty
variance (32\%); the factor most often named as \emph{over-weighted} is the
favourite's squad quality (40\%) and attacking firepower (38\%). These yield
concrete, minable priors for future systems: down-weight a favourite's paper
talent against a disciplined low block, and explicitly price the
draw$\rightarrow$shootout path in knockout ties.

\section{Discussion and Impact}
The benchmark separates three capabilities that a single-answer score conflates.
On prediction the 2026 agents are interchangeable and no better than the market;
on \emph{decision} and \emph{self-knowledge} they differ by large, stable
margins. For practitioners the lesson is that selecting an LLM for a
consequential forecasting-and-acting pipeline should weigh calibration, staking
discipline, and honest self-assessment, not headline accuracy. Because the
protocol is a thin wrapper around a consumer assistant, WC2026-Agents is a
template: the same search--act--reflect loop and market settlement apply to any
scheduled, market-priced event stream (other tournaments, elections, product
launches), enabling live, rolling, contamination-free evaluation and longitudinal
tracking of model releases. We release the loop, lexicon, and settlement code to
that end.

Two results deserve emphasis for benchmark design. First, the convergence of
four independently built models cautions against reading a single leaderboard
number as a capability signal: when models share a retrieval surface they also
share errors, so an ensemble inherits rather than cancels the common bias --- a
limit on the ``wisdom of the silicon crowd''~\cite{schoenegger2024wisdom} when
the crowd reads the same web. Second, that no agent beats a vig-bearing market is
the sharpest available null for an over-eager ``LLMs can predict the future''
narrative: the market is not a weak baseline but a
near-efficient~\cite{fama1970efficient,strumbelj2014odds} aggregator that the
agents match but do not exceed, so a benchmark that omits it would mistake
market-level performance for super-human forecasting. Surfacing such failure
modes is precisely the role of a datasets-and-benchmarks contribution.

\section{Using WC2026-Agents}
\label{sec:using}
The release supports several lines of work out of the box, each a single join on
the tidy 104-match table. \textbf{(i) Calibration and uncertainty.} 416
forecasts with confidences and outcomes, plus a market reference, form a
contamination-free testbed for LLM (mis)calibration and confidence
elicitation~\cite{tian2023calibration,xiong2024uncertainty}. \textbf{(ii) Agentic
decision-making.} Paired probabilities, stakes, and real odds let researchers
score arbitrary staking policies (flat, Kelly, learned) and quantify the value of
following versus fading a price --- a decision layer absent from prediction-only
benchmarks~\cite{karger2024forecastbench}. \textbf{(iii) Self-knowledge.} 414
paired reflections let one test whether models know when they were
wrong~\cite{kadavath2022know,huang2024selfcorrect} and whether self-reports
predict error, using our \textsf{outcome\_vs\_prediction} labels against realized
correctness. \textbf{(iv) Reasoning and evidence use.} Verbatim reasoning with
cited URLs supports factor-attribution and source-use studies via the released
lexicon. \textbf{(v) Temporal robustness and longitudinal tracking.} Because every
target post-dates training, the data probes genuine forecasting rather than
recall; the thin search--act--reflect protocol re-runs on any future tournament,
enabling live, rolling evaluation of successive model releases against a moving,
uncontaminated target.

\section{Ethics}
All betting in this study is virtual; the dataset is a measurement instrument,
not gambling advice, and we report profit only to score decision quality. No
human subjects or personal data are involved: the records are model outputs and
public fixtures, scores, and odds. Odds come from public preview articles with
source URLs; we take no book's proprietary feed. A foreseeable misuse ---
treating any agent's edge as an actionable betting strategy --- is unsupported
by our data, which show no agent beats the market. We disclose that generative
AI assisted the parsing and drafting of this work; all findings were verified
against the released data by the authors.

\section{Limitations}
This is a single tournament (104 matches); per-agent differences have wide
intervals and we stress patterns over precise rankings. Agents were queried
through consumer interfaces with tools enabled, so retrieval and decoding are not
fully controlled and providers may update models mid-window --- we hold the
user-visible model fixed and record it. Odds are opening lines from mostly one
book, not closing consensus. Reflections are the model's own free text, coded by
a transparent but lexical scheme. These are the natural costs of measuring
deployed agents on real, unrepeatable events, and the released pipeline makes
every choice auditable.

\section{Conclusion}
WC2026-Agents turns a future World Cup into a contamination-free benchmark of
LLMs as forecasting agents, paired with the betting market as a fifth,
economically grounded competitor. Its reference evaluation shows that frontier
agents predict alike and cannot beat the market, yet diverge sharply in
calibration, betting discipline, and the honesty of their self-reflection --- the
axes on which they actually differ. We release the data, odds, and evaluation
suite to support live, rolling evaluation of agentic forecasting.

\bibliographystyle{ACM-Reference-Format}
\bibliography{references}

\appendix
\section{Reproducibility}
One command (\texttt{python src/run\_all.py}) rebuilds every processed table,
analysis CSV, and figure from the raw transcripts and odds; the paper's numbers
are read directly from those outputs. The repository documents the schema,
the 13-category reasoning lexicon, per-match ground truth with penalty and
advancement fields, and all parse-issue flags. Randomness is not used in
evaluation. The four-way score cross-check and the odds/citation validation are
part of the build and re-run on every invocation.

\section{Per-stage metrics and calibration curves}
\begin{figure}[H]
  \centering
  \includegraphics[width=\columnwidth]{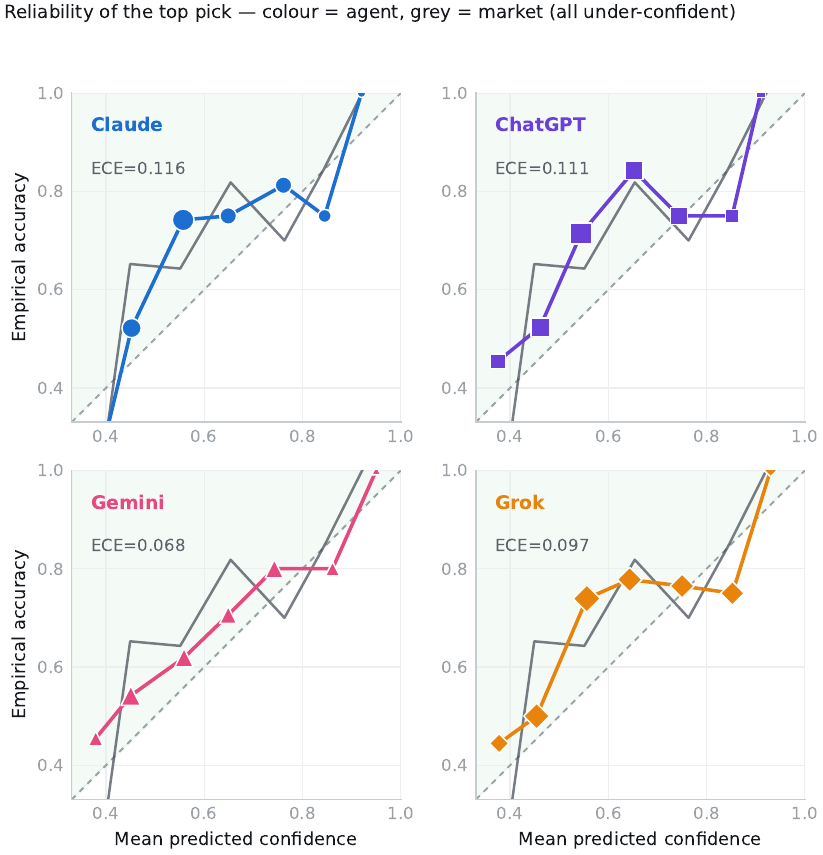}
  \caption{Reliability of each agent's top pick (App.). All lie in the shaded
  (under-confident) region; marker area $\propto$ bin count. ECE values also
  appear in Table~\ref{tab:t1}.}
  \label{fig:calibration}
\end{figure}
Table~\ref{tab:stage} breaks T1 down by phase. Knockout matches are markedly more
predictable for every competitor (accuracy $.719$ vs.\ $\sim.65$; lower Brier),
because seeding concentrates mismatches; the four agents are again
indistinguishable within each phase, and the market again leads on Brier. The
full 13$\times$4 factor table, the reliability bins behind
Fig.~\ref{fig:calibration}, the contrarian breakdown behind
Fig.~\ref{fig:contrarian}, and the match-level upset details are released as CSVs
under \texttt{data/analysis/} and regenerated by \texttt{src/run\_all.py}.

\begin{table}[h]
\caption{Task T1 by phase (accuracy / Brier / log-loss). Group ($n{=}72$) and
knockout ($n{=}32$).}
\label{tab:stage}
\small
\resizebox{\columnwidth}{!}{%
\begin{tabular}{lccc@{\hskip 1.2em}ccc}
\toprule
& \multicolumn{3}{c}{Group} & \multicolumn{3}{c}{Knockout} \\
\cmidrule(r){2-4}\cmidrule(l){5-7}
& Acc & Brier & LogL & Acc & Brier & LogL \\
\midrule
Claude  & .639 & .489 & .828 & .719 & .428 & .756 \\
ChatGPT & .667 & .493 & .838 & .719 & .429 & .759 \\
Gemini  & .625 & .485 & .820 & .719 & .477 & .822 \\
Grok    & .667 & .491 & .826 & .719 & .425 & .752 \\
\midrule
Market  & .667 & .486 & .827 & .719 & .431 & .761 \\
\bottomrule
\end{tabular}}
\end{table}

\end{document}